\def\year{2020}\relax
\documentclass[letterpaper]{article} 
\usepackage{aaai20}  
\usepackage{times}  
\usepackage{helvet} 
\usepackage{courier}  
\usepackage[hyphens]{url}  
\usepackage{graphicx} 
\usepackage{graphicx}  
\usepackage{makecell}
\usepackage{multirow}
\usepackage{amssymb}

\title{Co-Attention Hierarchical Network: \\ Generating Coherent Long Distractors for Reading Comprehension}
\author{
Xiaorui Zhou,\textsuperscript{\rm 1} 
Senlin Luo,\textsuperscript{\rm 1} 
Yunfang Wu\textsuperscript{\rm 2}\thanks{ Corresponding author.}\\ 
\textsuperscript{\rm 1}School of Information and Electronics, Beijing Institute of Technology \\
\textsuperscript{\rm 2}MOE Key Lab of Computational Linguistics, School of EECS, Peking University \\ 
zhouxr@bit.edu.cn, luosenlin@bit.edu.cn, wuyf@pku.edu.cn 
}
\begin{document}

\maketitle

\begin{abstract}
In reading comprehension, generating sentence-level distractors is a significant task, which requires a deep understanding of the article and question. The traditional entity-centered methods can only generate word-level or phrase-level distractors. Although recently proposed neural-based methods like sequence-to-sequence (Seq2Seq) model show great potential in generating creative text, the previous neural methods for distractor generation ignore two important aspects. First, they didn't model the interactions between the article and question, making the generated distractors tend to be too general or not relevant to question context. Second, they didn't emphasize the relationship between the distractor and article, making the generated distractors not semantically relevant to the article and thus fail to form a set of meaningful options. To solve the first problem, we propose a co-attention enhanced hierarchical architecture to better capture the interactions between the article and question, thus guide the decoder to generate more coherent distractors. To alleviate the second problem, we add an additional semantic similarity loss to push the generated distractors more relevant to the article. Experimental results show that our model outperforms several strong baselines on automatic metrics, achieving state-of-the-art performance. Further human evaluation indicates that our generated distractors are more coherent and more educative compared with those distractors generated by baselines. 
\end{abstract}

\section{Introduction}
\noindent Reading comprehension (RC) is an advanced cognitive activity of human beings, which involves interpretation of the text and making complex inferences \cite{DBLP:journals/corr/ChenBM16a}. The most popular form of assessment for reading comprehension is Multiple Choice Question (MCQ), since MCQs have many advantages including quick evaluation, less testing time, consistent scoring, and automatic evaluation \cite{article}. Besides the article itself, a MCQ consists of three elements: (i) \textit{stem}, the question body; (ii) \textit{key}, the correct answer; (iii) \textit{distractors}, alternative answers used to distract examinees from the correct answer. The effectiveness of MCQs depends not only on the validity of the question and the correct answer, but also on the quality of distactors \cite{goodrich1977distractor}. Among all methods for creating good MCQs, finding reasonable distractors is crucial and usually the most time-consuming \cite{liang2018distractor}.

\begin{figure}[t]
\centering
\includegraphics[width=0.95\columnwidth]{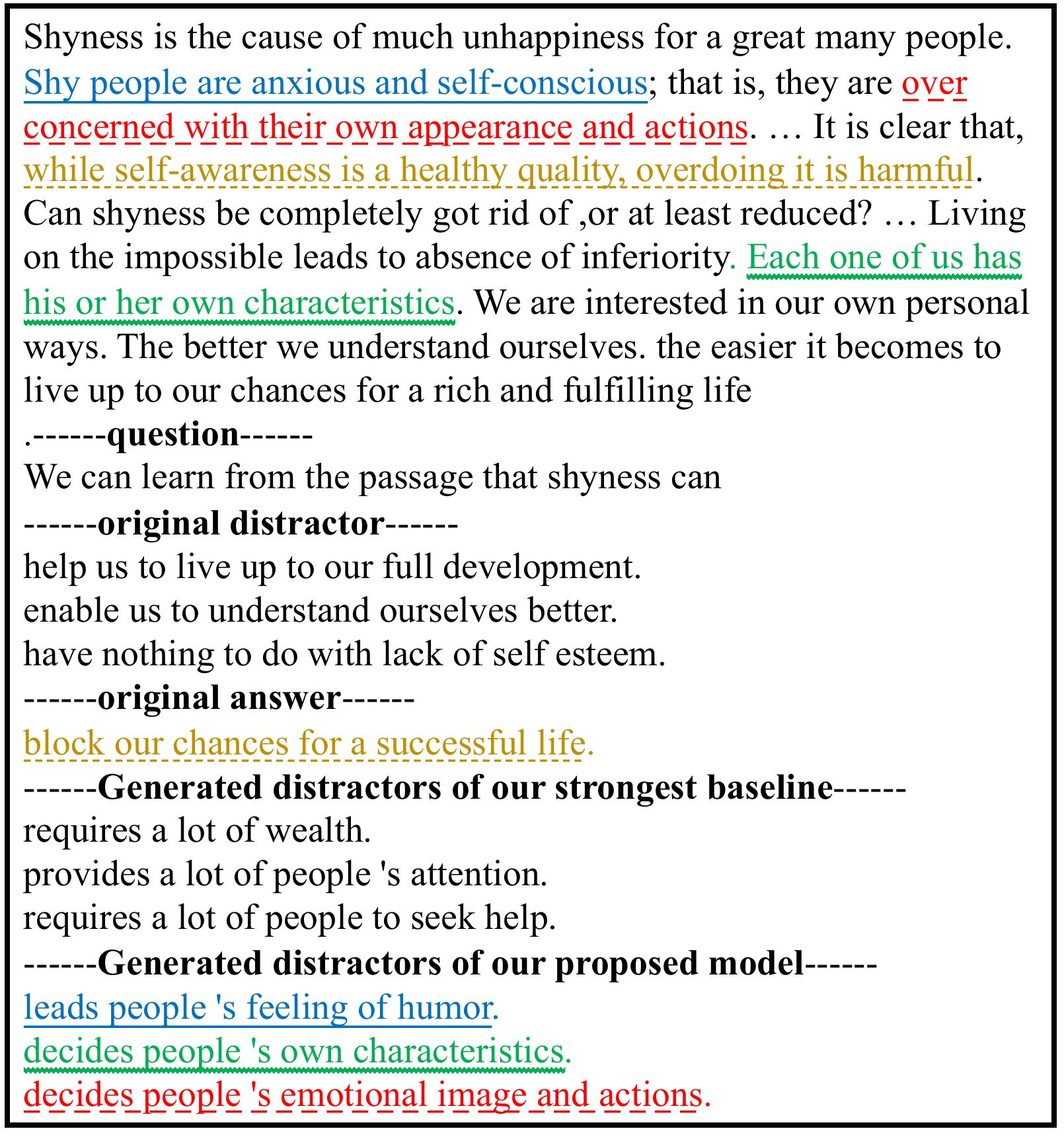} 
\caption{An example from our dataset, with the generated distractors of our strongest baseline and our proposed model, we use colors and underlines to indicate the semantic connection between MCQ segments and the article text.}
\label{fig1}
\end{figure}

In real examinations, a question for reading comprehension usually requires summarizing the article, or making inferences about a certain detail in the article. Figure 1 shows an example of MCQ from the RACE \cite{lai2017race} dataset, which records real English exams for middle and high school Chinese students. Different from the SQuAD \cite{RajpurkarZLL16} dataset which is widely used in RC research, in RACE the answer is a newly-generated sequence with the length of a sentence other than a text span extracted from the original text. Accordingly, the distractor should also be a sequence of words that is fluent and grammatical. More importantly, the distractor should be coherent with the question, and semantically relevant to the article. We call this type of distractor \textit{long distractor} to distinguish it from the word-level or phrase-level distractor in \textit{fill-in-the-blank} \cite{liang2017distractor} or \textit{cloze} MCQs. In this paper, we investigate the task of generating coherent long distractors for reading comprehension MCQs.

Traditionally, distractor generation is a component of an automatic MCQ generation system, and seldom has been taken out as a separate task. The process of generating MCQ \cite{article} usually consists of: (i) sentence selection: select a sentence that contains a questionable fact as a candidate for generating MCQs. (ii) Key selection: determine which word, n-gram, or phrase in the selected sentence should be blanked out. (iii) Question formation: transform a declarative sentence into the interrogative form. (iv) Distractor generation: generate distractors that are able to confuse the examinees. Approaches for generating distractors may utilize \textit{WordNet} \cite{miller1995wordnet}  to find synonyms or other related words as distractors, or use an existing domain-specific ontology to find related phrases \cite{stasaski2017multiple,araki2016generating},  or adopt other similarity-based methods like word embedding similarities \cite{guo2016questimator,kumar2015revup}, and co-occurrence likelihoods \cite{hill2016automatic}, etc. As we can see, the traditional MCQ generation is in a pipeline fashion, which requires human-designed features and external knowledge bases. Moreover, all the above mentioned approaches are based on entity relations, which can only generate word-level or phrase-level distractors and are not able to generate long distractors. 

Recently, deep neural models like Seq2Seq \cite{sutskever2014sequence} have achieved great success in a lot of Natural Language Processing (NLP) tasks, including machine translation, text summarization, headline generation and story generation. The recent work \cite{gao2019generating} employs a hierarchical encoder-decoder framework and proposes a  \textit{static attention} to notice the sentences related to the question and penalize the sentence related to the answer to generate long distactors on RACE dataset, and the proposed model outperforms several baselines, achieving a BLEU-4 score of 6.47 for the first distrator.

However, there is still much room for improvement in generating distractors. First, all previous proposed neural-based models adopt a simple Seq2Seq structure to build a direct mapping from article to distractor, which fails to model the interactions between the article and question, so the generated distractors tend to be too general that is not relevant to the theme of the article or not consistent with the question context. Figure 1 shows an example of distractors generated by our strongest baseline, which contain key words irrelevant to the article and question. As has been proved previously in the RC task \cite{seo2016bidirectional,XiongZS16}, capturing the complex interactions between article and question can improve performance in selecting correct answer. Second, previous proposed methods did not emphasize the relevance between the generated distractor and article. As a result, some of generated distrators are semantically far away from the article, thus fail to form a set of meaningful and educative options.

In this paper, we propose a Co-attention Hierarchical Network to generate distractors. The basic framework is a hierarchical encoder-decoder network with dynamic attention, which first obtains word-level hidden representations and then based on them to obtain the sentence-level representations. The dynamic attention combines word-level and sentence-level attention at each decoding time step. Based on this framework, we propose to incorporate co-attention mechanism, i.e. article-to-question and question-to-article, to allow the encoder better capture the rich interactions between the article and question. Further, we introduce a semantic similarity loss between the generated distractor and article into the original loss function, to guide the decoder to generate distractors that are more relevant to the article content.
  
We conduct extensive experiments on the challenging RACE dataset. Comparing with different approaches, our full model obtains the best results across most metrics (BLEU and ROUGE) for all three distractors. It outperforms the existing best method \cite{gao2019generating}, achieving a new state-of-the-art performance of BLEU-4 7.01 for the first distractor. The ablation study validates the effectiveness of our two proposed components. Further human evaluation demonstrates that our model can generate better quality distractors that are more consistent with the article and have stronger distracting ability.

\section{Proposed Framework}
\subsection{Notations and Task Definition}
\noindent For each sample in our dataset, we have an article that contains \(k\) sentences \(T = (s_1, s_2,\ldots,s_k)\) and each sentence \(s_i = (w_{i,1},w_{i,2},\ldots,w_{i,l})\) is a word sequence where \(l\) denotes the length of it. We also have a question \(Q = (q_1,q_2,\ldots,q_m)\) for each sample, where \(m\) denotes the length of the question. Our task is to generate a wrong option (distractor) \(D = (d_1,d_2,\ldots,d_z)\) where \(z\) is the distractor sequence length. 

Formally, we define the the distractor generation (DG) task as generating the best wrong option in reading comprehension, which is the conditional log-likelihood of the predicted distractor \(D\), given the article \(T\) and question \(Q\), such that:
\begin{equation}
    \bar D = \mathop{\arg\max}_{D} \big[\log \textup{P}(D|T,Q)\big].
\end{equation}

\begin{figure*}[t]
\centering
\includegraphics[width=0.95\textwidth]{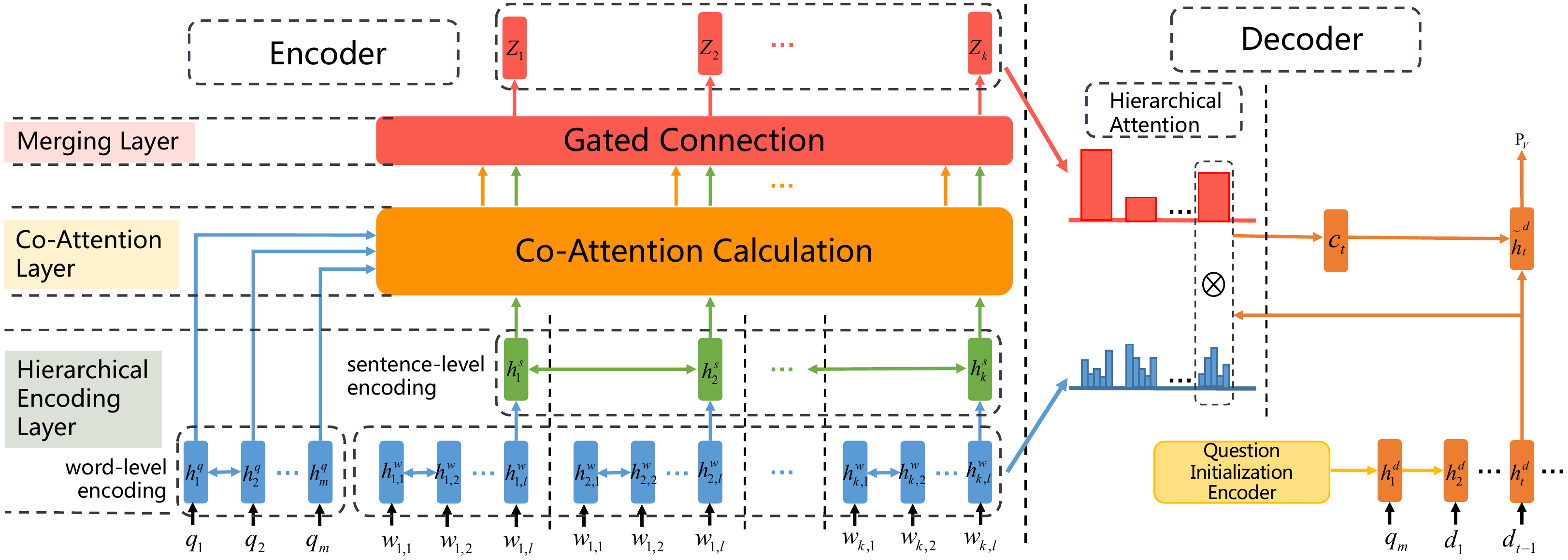} 
\caption{An overview of our proposed Co-Attention Hierarchical Network (Better viewed in color)}
\label{fig2}
\end{figure*}

\subsection{Model Overview}
\noindent In this paper, we propose a co-attention hierarchical network to generate distractors, as shown in Figure 2.

The encoder consists of three layers: 1) \textbf{Hierarchical Encoding Layer} maps input word embeddings to their word-level and sentence-level hidden representations. 2) \textbf{Co-Attention Layer} couples the question and article representations, and produces a set of question-aware feature vectors for each sentence in the article. 3) \textbf{Merging Layer} then merges the sentence-level representations and question-aware feature vectors to get the final sentence representations.

In the decoding phase, the word-level hidden representations and the final sentence representations are referenced at every decoding time step to calculate a hierarchical attention score. We employ a language model to compress the question into a fixed-length vector to initialize the decoder state, to make the distractor grammatically consistent with the question.

Moreover, we add a semantic similarity loss into the standard loss function, to enable the generated distractor more related to the article content.

\subsection{Encoding Article and Question}

\subsubsection{Hierarchical Article Encoder.} For each sentence \(s_i = (w_{i,1}, w_{i,2},\ldots,w_{i,l})\) in the article, we use a bidirectional LSTM \cite{hochreiter1997long} (denoted as \(\textup{LSTM}_{enc}^{w}\)) with hidden size \(r\) to encode this sequence into hidden representations 
\begin{equation}
    h_{i,t}^w = \textup{LSTM}_{enc}^{w}(h_{i,t-1}^{w},w_{i,t}).
\end{equation}
The vector output at the ending time step of this sequence is used as the embedding to represent the entire sentence:  
\begin{equation}
    e_i = h_{end_s}^w = h_{i,l}^{w}.
\end{equation}

In order to build representation \(e_T\) for the current article \(T\), another layer of LSTM (denoted as \(\textup{LSTM}_{enc}^{s}\)) with hidden size \(r\) is placed on top of all sentence embeddings, computing representations sequentially for each time step:
\begin{equation}
    h_t^s = \textup{LSTM}_{enc}^{s}(h_{t-1}^{s},e_{t}).
\end{equation}
Similarly, we use the final time step sentence-level representation \(h_{end_T}^s\) to represent the entire document:
\begin{equation}
    e_T = h_{end_T}^s = h_k^s.
\end{equation}
Let \(\textbf{H}^{*} \in\mathbb{R}^{r \times k}\) to denote the sentence-level representations of the article, where \(\textbf{H}_{:t}^{*} = h_t^s\).

By utilizing this hierarchical structure, we decompose a long document into a two-level connection of relatively short sequences, avoiding directly encoding the whole document through a single LSTM.

\subsubsection{Question Encoder.} We use a bidirectional LSTM to encode the question sequence \((q_1,q_2,\ldots,q_m)\) into hidden representation. In our implementation, we share the same LSTM with the word-level LSTM used in article encoding, so
\begin{equation}
    h_t^q = \textup{LSTM}_{enc}^{w}(h_{t-1}^q,q_t).
\end{equation}
We use \(\textbf{U}^{*} \in\mathbb{R}^{r \times m}\) to denote all word-level representations of the question, where \(\textbf{U}_{:t}^{*} = h_t^q\).

\subsection{Co-Attention between Article and Question}
\noindent To model the complex interactions between the article and question, we adopt a co-attention mechanism \cite{seo2016bidirectional}, which is computed in two directions: from article to question as well as from question to article. Specifically, these two types of attention are calculated between sentence-level representations of the article \(\textbf{H}^{*}\) and the word-level representations of the question \(\textbf{U}^{*}\).

First, we let \(\textbf{H}^{*}\) and \(\textbf{U}^{*}\) go through a \textit{dimension transformation} layer to shrink the dimension of these two representations,
\begin{equation}
    \textbf{H} = \textup{w}_{d}\textbf{H}^{*}+\textup{b}_{d} \in\mathbb{R}^{r/4 \times k},
\end{equation}
\begin{equation}
    \textbf{U} = \textup{w}_{d}\textbf{U}^{*}+\textup{b}_{d} \in\mathbb{R}^{r/4 \times m}.
\end{equation}
\(\textup{w}_{d}\) and \(\textup{b}_{d}\) are trainable parameters, here we must assume that the hidden size \(r\) is divisible by 4. 

Next, both the two directions of co-attention, which will be discussed below, are derived from a shared similarity matrix, \(\textbf{S}\in\mathbb{R}^{k\times m}\), between the transformed sentence representations (\textbf{H}) and the transformed question representations (\textbf{U}), where \(\textbf{S}_{ij}\) indicates the similarity between \textit{i}-th article sentence and \textit{j}-th question word. The similarity matrix is computed by
\begin{equation}
    \textbf{S}_{ij} = \phi(\textbf{H}_{:i},\textbf{U}_{:j}) \in\mathbb{R},
\end{equation}
where \(\phi\) is a trainable scalar function that encodes the similarity between its two input vectors, \(\textup{\textbf{H}}_{:i}\) is \(i\)-th column vector of \(\textup{\textbf{H}}\), and \(\textup{\textbf{U}}_{:j}\) is \(j\)-th column vector of \(\textup{\textbf{U}}\), We set: 
\begin{equation}
    \phi(\textbf{h}, \textbf{u}) = \textup{w}_{s}^{\top}[\textup{\textbf{h}};\textup{\textbf{u}};\textup{\textbf{h}}\circ \textup{\textbf{u}}],
\end{equation}
where \(\textup{w}_{s} \in\mathbb{R}^{3r/4}\) is a trainable weight vector, \(\circ\) is elementwise multiplication, \([;]\) is vector concatenation across row, and implicit multiplication is matrix multiplication.

Then, the similarity matrix is normalized for each column to produce the attention weights \(\textbf{S}^{\textup{Q}} \in\mathbb{R}^{m \times k}\) across the question words for each sentence in the article, and normalized for each row to produce the attention weights \(\textbf{S}^{\textup{T}} \in\mathbb{R}^{m \times k}\) across the article sentences for each word in the question:
\begin{equation}
    \textup{\textbf{S}}_{:j}^{\textup{Q}} = \textup{softmax}(\textup{\textbf{S}}_{:j}), \forall j;
\end{equation}
\begin{equation}
   \textup{\textbf{S}}_{i:}^{\textup{T}} = \textup{softmax}(\textup{\textbf{S}}_{i:}), \forall i.
\end{equation}

\subsubsection{Article-to-question Attention.} Article-to-question attention (A2Q) signifies which question words are most relevant to each article sentence. So for \(j\)-th sentence in the article, we sum over all question representations \(\textup{\textbf{U}}_{:i} \in \mathbb{R}^{r/4} , \forall i\) according to their normalized attention weights with that sentence \(\textup{\textbf{S}}_{ij}^{\textup{Q}} \in\mathbb{R}\). Subsequently, each attended question vector is computed by  \(\tilde{\textup{\textbf{U}}}_{:j} = \sum_{i}{\textup{\textbf{S}}_{ij}^{\textup{Q}}\textup{\textbf{U}}_{:i}}\). Hence \(\tilde{\textup{\textbf{U}}}\) is a \(r/4\)-by-\(k\) matrix containing the attended question vectors for the entire article sentences.

\subsubsection{Question-to-article Attention.} Question-to-article (Q2A) attention signifies which article sentences have the closest similarity to each of the question words and are hence critical for locating information that is most relevant for answering that question. We obtain the attended sentence vector on the article words by 
\begin{equation}
    \tilde{\textup{\textbf{H}}} = \textup{\textbf{H}} (\textup{\textbf{S}}^{\textup{T}})^{\top}\textup{\textbf{S}}^{\textup{Q}} \in\mathbb{R}^{r/4\times k},
\end{equation}
where \(\textup{\textbf{H}} (\textup{\textbf{S}}^{\textup{T}})^{\top} \in\mathbb{R}^{r/4 \times m}\) is the weighted sum of sentence representations for each question word, and \(\textup{\textbf{S}}^{\textup{Q}}\) is used to map the matrix to length \(k\).

Finally, the sentence representations and the attention vectors are combined together to yield \textup{\textbf{G}}, where each column vector can be considered as the question-aware representation of each article sentence. \textbf{G} is defined by 
\begin{equation}
    \textbf{G}_{:t} = \psi(\textbf{H}_{:t},\tilde{\textbf{U}}_{:t},\tilde{\textbf{H}}_{:t}) \in\mathbb{R}^{d_G},
\end{equation}
where \(\textbf{G}_{:t}\) is the \(t\)-th column vector (corresponding to \(t\)-th article sentence), \(\psi\) is a trainable vector function that fuses its (three) input vectors, and \(d_G\) is the output dimension of the \(\psi\) function. While the \(\psi\) function can be an arbitrary trainable neural network, in our experiments we adopt a simple concatenation as following:
\begin{equation}
    \psi(\textbf{h},\tilde{\textbf{u}},\tilde{\textbf{h}}) = [\textbf{h};\tilde{\textbf{u}};\textbf{h}\circ\tilde{\textbf{u}};\textbf{h}\circ\tilde{\textbf{h}}] \in \mathbb{R}^{r\times k}.
\end{equation}

\subsection{Merging Sentence Representation}
\noindent We then go through a gated connection layer to merge the sentence contextual representations and the question-aware representations
\begin{equation}
    \textup{\textbf{g}} = \sigma(\textbf{\textup{G}}) \in\mathbb{R}^{r \times k},
\end{equation}
\begin{equation}
    \textbf{\textup{Z}} = \textup{\textbf{g}} \circ \textbf{\textup{H}}^{*} + (1-\textup{\textbf{g}}) \circ \textbf{\textup{G}}.
\end{equation}
Where \(\sigma\) is Sigmoid function, and \textbf{Z} is the final representation of sentence-level hidden states. This new representation of sentence contains both sentence-level contextual information and the article question co-attention, thus is more likely to capture the article content that is more related to the given question.

\subsection{Hierarchical Attention Decoder}
\noindent We use another uni-directional LSTM as the decoder to generate distractor. 
\subsubsection{Question Initialization.} Unlike standard Seq2Seq generation task like machine translation, in which both source and target sequence are complete sentences, our dataset contain near half of the questions that are not complete sentences (as shown in Figure 1 and Table 1). In order to handle this problem, we follow previous work to use a question-based initializer \cite{gao2019generating}.

We use a uni-directional LSTM to encode the question sequence \((q_1,q_2,\ldots,q_m)\) into hidden representations, and denote the hidden state of the final step as \(h^{q_{init}}_{m}\). Then \(h^{q_{init}}_{m}\) is used as the initial state the decoder. Moreover, instead of using \textit{begin-of-sentence} symbol, we use the last token in the question \(q_m\) as the initial input of the decoder.

\subsubsection{Hierarchical Attention.}We employ hierarchical attention \cite{ling2017coarse} to attend the article with different granularity. At each decoding time-step, we parallelly calculate both the sentence-level attention weight \(\beta\) and word-level attention \(\alpha\) by
\begin{equation}
    \beta_i = {\textup{\textbf{Z}}_{:i}}^{\top}\textup{W}_{d_1}h_t^d,\quad \alpha_{i,j} = {h_{i,j}^w}^{\top}\textup{W}_{d_2}h_t^d,
\end{equation}
\begin{equation}
    \gamma_{i,j} = \frac{\alpha_{i,j}\beta_i}{\sum_{i,j}{\alpha_{i,j}\beta_i}}.
\end{equation}
Where \(\textup{W}_{d_1}\) and \(\textup{W}_{d_2}\) are trainable parameters. The sentence-level attention determines how much each sentence should contribute to the generation at the current time-step, while the word-level attention determines how to distribute the attention over words in each sentence.

Then the context vector \(\textbf{c}_t\) is derived as a combination of all word-level representations reweighted by the combined attention \(\gamma\):
\begin{equation}
    \textbf{c}_t = \sum_{i,j}{\gamma_{i,j}h_{i,j}^w}.
\end{equation}
And the attentional vector is calculated as:
\begin{equation}
    \tilde{h}_t^d = \textup{tanh}(\textup{W}_{\tilde{h}}[h_t^d;\textbf{c}_t]).
\end{equation}
Finally, the predicted probability distribution over the vocabulary \(V\) at the current step is computed as:
\begin{equation}
    \textup{P}_V = \textup{softmax}(\textup{W}_V\tilde{h}_t^d + \textup{b}_V),
\end{equation}
where \(\textup{W}_{\tilde{h}}\), \(\textup{W}_V\) and \(\textup{b}_V\) are learnable parameters.

\subsection{Semantic Similarity Loss}
\noindent We assume that in order to form a set of meaningful and educative options, each distractor should be semantically relevant to the article. So we incorporate an additional semantic similarity loss into the original loss function.

In order to calculate the semantic similarity score, we should first obtain the semantic representation vectors of the generated distractor. Previous work has proved that a simple subtraction between LSTM hidden states can represent segment sequence effectively \cite{wang2016graph}. So the distractor representation \(e_D\) is computed by:
\begin{equation}
    e_D = s_M - e_T,
\end{equation}
where \(s_M\) denotes the decoder last hidden state, and \(e_T\) is the sentence-level encoder's last hidden state, which is also the representation of the article. 

Then we compute the cosine similarity to measure the semantic relevance between distractor and article, which is represented with a dot product and magnitude:
\begin{equation}
    cos(e_D, e_T) = \frac{e_D\cdot e_T}{\|e_D\| \cdot \|e_T\|}.
\end{equation}
Our training objective is to maximize the similarity score so that the generated distractor have high semantic relevance with the article. Previous work in text summarization has proved that this cosine similarity loss can improve the semantic relevance of the source text and the generated summary  \cite{ma2017improving}.

Finally, the model is trained to minimize the total loss:
\begin{equation}
    \mathcal{L} = -\sum_{d \in V}\log \textup{\textup{P}}(d|T,Q; \Theta) - \lambda cos(e_D, e_T),
\end{equation}
where \(\lambda\) is a hyperparameter to balance two loss fucntions.

\section{Experimental Settings}
\subsection{Dataset}
\noindent We conduct extensive experiments on the RACE dataset, which was collected from the English exams for middle and high school Chinese students. It contains \(27,933\) articles with \(97,687\) questions, which are designed by human experts for education purpose, making RACE an ideal dataset for training model to generate questions and distrators.

Instead of using original RACE dataset which contains samples that are not suitable for sequence generation, we use the dataset processed by previous work \cite{gao2019generating} under the following two conditions: 1) Filter out the distractors that are semantic irrelevant to the article context. 2) Remove the questions which require to fill in the options at the beginning or in the middle of the questions. Table 1 shows the statistics of our dataset.

\subsection{Baselines and Evaluation Metrics}
\noindent We compare our model with the following baselines.
\begin{itemize}
    \item \textbf{Seq2Seq:} The standard sequence-to-sequence structure with global attention mechanism \cite{luong2015effective}. The encoder take the whole article sequence as input, and use a single LSTM to encoder this sequence.
    \item \textbf{HRED:} Vanilla hierarchical structure as described before. It consists of A hierarchical encoder that is able to encode document-level input text in a way that preserve semantic coherence \cite{li2015hierarchical}, and a decoder that utilize hierarchical attention. This structure has been proved effective in text summarization \cite{ling2017coarse} and headline generation \cite{tan2017neural}.
    \item \textbf{HRED+copy} (\textbf{HCP})\textbf{:} Incorporating pointer-generator-network \cite{See_2017} with \textbf{HERD} to enable the decoder to directly copy words from the source text. Pointer-generator-network made a big improvement in text generation tasks like text summarization, so we consider \textbf{HCP} to be a strong baseline.
    \item \textbf{HRED+static\_attn} (\textbf{HSA}) \cite{gao2019generating}\textbf{:} A \textit{static attention} that penalize the correlation between the answer and generated distractors was proposed to modulate the hierarchical attention in \textbf{HERD} , this model achieved state-of-the-art performance previously on this task.
\end{itemize}

Following the previous work \cite{gao2019generating}, We adopt BLEU \cite{Papineni2002BLEU} and ROUGE \cite{lin2004rouge} to evaluate the performance of our models.

\begin{table}[t]
\centering
\resizebox{.95\columnwidth}{!}{
\smallskip\begin{tabular}{l c c c}
\Xhline{3\arrayrulewidth}
 & Train & Valid & Test\\
\hline
\# samples & 96501 & 12089 & 12284 \\
\% incomplete-sentence questions & 47.48 & 46.48 & 47.57 \\
Avg. article length & 347.21 & 344.78 & 347.66 \\
Avg. question length & 9.91 & 9.97 & 9.93 \\
Avg. distractor length & 8.50 & 8.50 & 8.54 \\
\Xhline{3\arrayrulewidth}
\end{tabular}
}
\caption{The statistics of our dataset.}\smallskip
\label{table1}
\end{table}

\begin{table*}[t]
\centering
\smallskip
\small
\begin{tabular}{c l c c c c c c c}
\Xhline{3\arrayrulewidth}
 & & BlEU-1 & BlEU-2 & BLEU-3 & BLEU-4 & ROUGE-1 & ROUGE-2 & ROUGE-L \\
\hline
\multirow{6}{*}{1st Distractor} & \(\textup{Seq2Seq}^{*}\) & 25.28 & 12.43 & 7.12 & 4.51 & 14.12 & 3.35 & 13.58 \\
  & HRED & 27.96 & 14.41 & 9.05 & 6.34 & 15.55 & 3.97 & 14.68 \\
  & HCP & 26.13 & 13.26 & 8.81 & 6.68 & 14.60 & 3.72 & 13.84 \\
  & \(\textup{HSA}^{*}\) & 27.32 & 14.69 & 9.29 & 6.47 & 15.69 & \textbf{4.42} & 15.12 \\
  & HSA & 28.18 & 14.57 & 9.19 & 6.43 & 15.74 & 4.02 & 14.89 \\
  & Our Model & \textbf{28.65} & \textbf{15.15} & \textbf{9.77} & \textbf{7.01} & \textbf{16.22} & 4.34 & \textbf{15.39} \\
\hline
\multirow{6}{*}{2st Distractor} & \(\textup{Seq2Seq}^{*}\) & 25.13 & 12.02 & 6.56 & 3.93 & 13.72 & 3.09 & 13.20 \\
  & HRED & 27.85 & 13.39 & 7.89 & 5.22 & 15.51 & 3.44 & 14.48 \\
  & HCP & 24.01 & 10.33 & 5.84 & 3.88 & 13.04 & 2.52 & 12.22 \\
  & \(\textup{HSA}^{*}\) & 26.56 & 13.14 & 7.58 & 4.85 & 14.72 & 3.52 & 14.15 \\
  & HSA & \textbf{27.85} & 13.41 & 7.87 & 5.17 & 15.35 & 3.40 & 14.41 \\
  & Our Model & 27.29 & \textbf{13.57} & \textbf{8.19} & \textbf{5.51} & \textbf{15.82} & \textbf{3.76} & \textbf{14.85} \\
\hline
\multirow{6}{*}{3st Distractor} & \(\textup{Seq2Seq}^{*}\) & 25.34 & 11.53 & 5.94 & 3.33 & 13.78 & 2.82 & 13.23 \\
  & HRED & 26.73 & 12.55 & 7.21 & 4.58 & 15.96 & \textbf{3.46} & 14.86 \\
  & HCP & 23.93 & 10.68 & 6.34 & 4.38 & 13.71 & 2.84 & 12.84 \\
  & \(\textup{HSA}^{*}\) & 26.92 & \textbf{12.88} & 7.12 & 4.32 & 14.97 & 3.41 & 14.36 \\
  & HSA & \textbf{26.93} & 12.62 & 7.25 & 4.59 & 15.80 & 3.35 & 14.72 \\
  & Our Model & 26.64 & 12.67 & \textbf{7.42} & \textbf{4.88} & \textbf{16.14} & 3.44 & \textbf{15.08} \\
\Xhline{3\arrayrulewidth}
\end{tabular}
\caption{Automatic evaluation results of different models. * symbol indicates the results are taken from the original paper of \cite{gao2019generating}. The best performing result for each metric is highlighted in boldface.}
\label{table2}
\end{table*}

\subsection{Implementation Details}
\noindent Our training set contains 100,116 distinct words and we keep the most frequent 50k tokens as vocabulary. Those tokens outside the vocabulary are replaced by the UNK symbol. The hidden unit size of all LSTMs is set to 600. The word-level encoder and sentence-level encoder are bidirectional LSTMs with their number of layer to be 2 and 1 respectively. The question initialization encoder and the decoder are 2 layers unidirectional LSTMs. We use the \texttt{GloVe.840B.300d} word embeddings and make them trainable. For optimization in training, we use stochastic gradient descent (SGD) as the optimizer and set the gradient norm upper bound to \(5.0\). We set minibatch size to 10 and the initial learning rate to \(1.0\) with a decay rate of 0.8. For hyperparameter \(\lambda\) of semantic similarity loss, we tested in a large range and set it to 0.0001.

During inference, we adopt beam search and set beam size to 10. We keep the 10 best candidate distractors in decending likelihood, and utilize a Jaccard distance of 0.5 to select three diverse distractors. Therefore, the Jaccard distance between distractor \(D_2\) and \(D_1\) is larger than 0.5, and the Jaccard distance between distractor \(D_3\) and both \(D_1\) and \(D_2\) is also greater than 0.5.

\section{Results and Analysis}
\subsection{Main Results}
The experimental results are shown in Table 2. We also list the results of \textbf{Seq2Seq} and \textbf{HSA} from the previous work \cite{gao2019generating} for reference. Our model achieves the best results across most metrics for all three distractors. As for the first distractor that is most important in our task, our model obtains a new state-of-the-art performance of 7.01 at BLEU-4 metric, which outperforms the existing best result \cite{gao2019generating} by 0.54 points. 

As shown in Table 2, there is a large performance gap between \textbf{Seq2Seq} and \textbf{HERD}, which reveals that the hierarchical structure is indeed effective for keeping semantic information of long sequential input. It also can be found that incorporating copy mechanism does not improve performance, we think one reason is that the copy mechanism mainly handle out-of-vocabulary (OOV) words problem, and our dataset is used for education purpose, there is only an OOV words ratio of 0.51\%  among all distractor's word occurrences.


\begin{table}[t]
\centering
\resizebox{.85\columnwidth}{!}{
\smallskip\begin{tabular}{l|c c c c}
\Xhline{3\arrayrulewidth}
 & BLEU-3 & BLEU-4 & ROUGE-L\\
\hline
HRED & 9.05 & 6.34 & 14.68 \\
+ SSL & 9.51 & 6.88 & 14.59 \\
+ Co-Attn & 9.66 & 6.85 & 15.17 \\
+ Co-Attn - Merging & 9.07 & 6.37 & 14.49 \\
Our Model & \textbf{9.74} & \textbf{7.01} & \textbf{15.19} \\
\Xhline{3\arrayrulewidth}
\end{tabular}
}
\caption{Ablation study of our model. "+SSL" means adding the semantic similarity loss, and "+Co-Attn" means adding the co-attention network. "-Merging" means getting rid of the merging layer. Here we only list results of the first distractor.}\smallskip
\label{table3}
\end{table}

\subsection{Ablation Study}
\noindent We compare the results of our full model with its ablated variants to analyze the relative contributions of each component. The results are shown in Table 3. It indicates that both our proposed co-attention architecture and the semantic similarity loss improve the performance over the basic \textbf{HRED} model obviously, and combining them together helps our full model achieve state-of-the-art performance. 

We also verified the necessity of the gated connection layer. By getting rid of merging layer and using question-aware representation \textbf{G} as our sentence-level representation, the model performance actually decrease. This shows it's necessary to keep the original sentence-level representation \(\textbf{\textup{H}}^{*}\).

\begin{table}[t]
\centering
\resizebox{.95\columnwidth}{!}{
\smallskip\begin{tabular}{l|c c c}
\Xhline{3\arrayrulewidth}
 & Fluency & Coherence & Distracting Ability \\
\hline
HRED & 7.81 & \textbf{7.38} & 4.93 \\
HSA & 7.93 & 6.86 & 4.28 \\
Our Model & \textbf{8.04} & 7.32 & \textbf{5.23} \\
\Xhline{3\arrayrulewidth}
\end{tabular}
}
\caption{Results of human evaluation.}\smallskip
\label{table4}
\end{table}

\begin{figure*}[t]
\centering
\includegraphics[width=.95\textwidth]{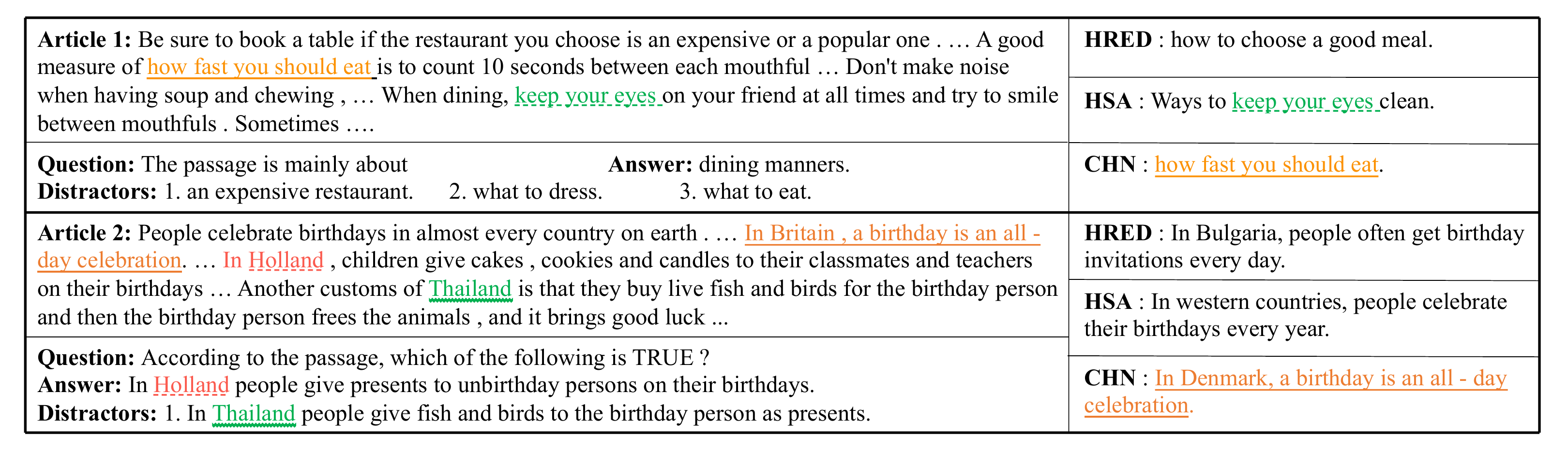} 
\caption{Samples of distractors generated by our model (CHN) and two other competitive models, HRED \cite{li2015hierarchical,ling2017coarse} and HSA \cite{gao2019generating}. We use colors and underlines to indicate the semantic connection between the distractor segments and article text.}
\label{fig3}
\end{figure*}

\subsection{Human Evaluation}
\noindent We also conduct a human evaluation to evaluate the generated distractors of our different models. We design three metrics to evaluate the quality of generated distractors. For all the metrics, we ask the annotator to score the distractors with three gears, the scores are then projected to 0 - 10. We employ three annotators to evaluate the distractors generated by our three most competitive models over the first 100 samples of the test set.
\begin{itemize}
    \item \textbf{Fluency:} This metric evaluates whether the distractor follows regular English grammar and whether the distractor accords with human logic and common sense.
    \item \textbf{Coherence:} This metric looks for key phrases in the distractors and measures whether these key phrases are relevant to the article and the question.
    \item \textbf{Distracting Ability:} This metric evaluates how likely a distractor candidate will be chosen as distractor in real examinations. This metric is designed to detect whether a distractor tries to mislead the examinees to an irrelevant topic, or in other words, distract by misleading.
\end{itemize}
The results are presented in Table 4. It is amazing to find that vanilla hierarchical structure model \textbf{HERD} actually yields quite competitive results. Our model get the highest scores in \textit{Fluency} and \textit{Distracting Ability} metrics, and a nearly best score in \textit{Coherence} metric. This shows that our model is able to generate more coherent and educative distractors. \textbf{HSA} gets a low score in \textit{Distracting Ability} metric, we hypothesis that this is because \textit{static attention} penalizes the correlation between the generated distractors and the correct answer, so the generated distractors tend to be semantically far away from the correct answer and less relevant to the article.

\subsection{Case Study}
\noindent In Figure 1, We show an example of distractors generated by our strongest baseline (\textbf{HSA}) and our model. This article is about shyness and the way to overcome it. Distractors generated by \textbf{HSA} contain words like \textit{wealth}, \textit{provides}, and \textit{seek help}, which are not relevant to the content of this article. So if examinees do not understand this article very well, these irrelevant distractors may confuse them more, leading them to wrongly choose these irrelevant distractors. Previous work has proved that distractors generated by \textbf{HSA} are most successful in confusing the annotators \cite{gao2019generating}, we hypothesis that part of this confusing ability comes from their misleading property, thus these distractors do not serve a good education purpose. We also present some general cases in Figure 3, the first article is about dining manners. Distractor generated by \textbf{HSA} shares some common words with the article, but it's semantically irrelevant to the article. And our model (\textbf{CHN}) generates more coherent distractor than \textbf{HRED}. The second article is about different customs of celebrating birthday in different countries. Distractor generated by \textbf{HSA} is too general to be meaningful, and the article did'nt mention or talk about \textit{birthday invitations}, so the distractor generated by \textbf{HRED} is not a good one. While distractor generated by our model only changed the country name in the original article sentence, examinees need to carefully check the article content to make a judgment. Therefore, it is a coherent and educative distractor.

\section{Conclusion}
\noindent In this paper, we present a Co-Attention Hierarchical Network to generate coherent long distractors for reading comprehension multiple choice questions. A co-attention enhanced hierarchical architecture is exploited to model the complex interactions between the article and question, guiding the decoder to generate more coherent and consistent distractors. Then a semantic similarity loss is incorporated into the original loss   
to push the generated distractors to be more relevant to the article content. Our model outperforms several strong baselines including the existing best model. The ablation study verifies the robustness of our model, and human evaluation shows our model is able to generate more coherent and educative distractos. For the future work, some interesting directions include exploring more complex co-attention structure and utilizing the information provided by the correct answer.

\section{Acknowledgments}
\noindent We thank Wenjie Zhou for his valuable comments and suggestions. This work is supported by the National Natural Science Foundation of China (61773026) and the Key Project of Natural Science Foundation of China (61936012).

\bibliographystyle{aaai}
\bibliography{AAAI-ZhouX.7388}

\begin{thebibliography}{}

\bibitem[\protect\citeauthoryear{Araki \bgroup et al\mbox.\egroup
  }{2016}]{araki2016generating}
Araki, J.; Rajagopal, D.; Sankaranarayanan, S.; Holm, S.; Yamakawa, Y.; and
  Mitamura, T.
\newblock 2016.
\newblock Generating questions and multiple-choice answers using semantic
  analysis of texts.
\newblock In {\em Proceedings of COLING 2016, the 26th International Conference
  on Computational Linguistics: Technical Papers},  1125--1136.

\bibitem[\protect\citeauthoryear{Chen, Bolton, and
  Manning}{2016}]{DBLP:journals/corr/ChenBM16a}
Chen, D.; Bolton, J.; and Manning, C.~D.
\newblock 2016.
\newblock A thorough examination of the cnn/daily mail reading comprehension
  task.
\newblock {\em CoRR} abs/1606.02858.

\bibitem[\protect\citeauthoryear{Gao \bgroup et al\mbox.\egroup
  }{2019}]{gao2019generating}
Gao, Y.; Bing, L.; Li, P.; King, I.; and Lyu, M.~R.
\newblock 2019.
\newblock Generating distractors for reading comprehension questions from real
  examinations.
\newblock In {\em Proceedings of the AAAI Conference on Artificial
  Intelligence}, volume~33,  6423--6430.

\bibitem[\protect\citeauthoryear{Goodrich}{1977}]{goodrich1977distractor}
Goodrich, H.~C.
\newblock 1977.
\newblock Distractor efficiency in foreign language testing.
\newblock {\em Tesol Quarterly}  69--78.

\bibitem[\protect\citeauthoryear{Guo \bgroup et al\mbox.\egroup
  }{2016}]{guo2016questimator}
Guo, Q.; Kulkarni, C.; Kittur, A.; Bigham, J.~P.; and Brunskill, E.
\newblock 2016.
\newblock Questimator: Generating knowledge assessments for arbitrary topics.
\newblock In {\em IJCAI-16: Proceedings of the AAAI Twenty-Fifth International
  Joint Conference on Artificial Intelligence}.

\bibitem[\protect\citeauthoryear{Hill and Simha}{2016}]{hill2016automatic}
Hill, J., and Simha, R.
\newblock 2016.
\newblock Automatic generation of context-based fill-in-the-blank exercises
  using co-occurrence likelihoods and google n-grams.
\newblock In {\em Proceedings of the 11th Workshop on Innovative Use of NLP for
  Building Educational Applications},  23--30.

\bibitem[\protect\citeauthoryear{Hochreiter and
  Schmidhuber}{1997}]{hochreiter1997long}
Hochreiter, S., and Schmidhuber, J.
\newblock 1997.
\newblock Long short-term memory.
\newblock {\em Neural computation} 9(8):1735--1780.

\bibitem[\protect\citeauthoryear{Kumar, Banchs, and
  D’Haro}{2015}]{kumar2015revup}
Kumar, G.; Banchs, R.; and D’Haro, L.~F.
\newblock 2015.
\newblock Revup: Automatic gap-fill question generation from educational texts.
\newblock In {\em Proceedings of the Tenth Workshop on Innovative Use of NLP
  for Building Educational Applications},  154--161.

\bibitem[\protect\citeauthoryear{Lai \bgroup et al\mbox.\egroup
  }{2017}]{lai2017race}
Lai, G.; Xie, Q.; Liu, H.; Yang, Y.; and Hovy, E.
\newblock 2017.
\newblock Race: Large-scale reading comprehension dataset from examinations.
\newblock {\em arXiv preprint arXiv:1704.04683}.

\bibitem[\protect\citeauthoryear{Li, Luong, and
  Jurafsky}{2015}]{li2015hierarchical}
Li, J.; Luong, M.-T.; and Jurafsky, D.
\newblock 2015.
\newblock A hierarchical neural autoencoder for paragraphs and documents.
\newblock {\em arXiv preprint arXiv:1506.01057}.

\bibitem[\protect\citeauthoryear{Liang \bgroup et al\mbox.\egroup
  }{2017}]{liang2017distractor}
Liang, C.; Yang, X.; Wham, D.; Pursel, B.; Passonneaur, R.; and Giles, C.~L.
\newblock 2017.
\newblock Distractor generation with generative adversarial nets for
  automatically creating fill-in-the-blank questions.
\newblock In {\em Proceedings of the Knowledge Capture Conference}, ~33.
\newblock ACM.

\bibitem[\protect\citeauthoryear{Liang \bgroup et al\mbox.\egroup
  }{2018}]{liang2018distractor}
Liang, C.; Yang, X.; Dave, N.; Wham, D.; Pursel, B.; and Giles, C.~L.
\newblock 2018.
\newblock Distractor generation for multiple choice questions using learning to
  rank.
\newblock In {\em Proceedings of the Thirteenth Workshop on Innovative Use of
  NLP for Building Educational Applications},  284--290.

\bibitem[\protect\citeauthoryear{Lin}{2004}]{lin2004rouge}
Lin, C.-Y.
\newblock 2004.
\newblock Rouge: A package for automatic evaluation of summaries.
\newblock In {\em Text summarization branches out},  74--81.

\bibitem[\protect\citeauthoryear{Ling and Rush}{2017}]{ling2017coarse}
Ling, J., and Rush, A.
\newblock 2017.
\newblock Coarse-to-fine attention models for document summarization.
\newblock In {\em Proceedings of the Workshop on New Frontiers in
  Summarization},  33--42.
\newblock Copenhagen, Denmark: Association for Computational Linguistics.

\bibitem[\protect\citeauthoryear{Luong, Pham, and
  Manning}{2015}]{luong2015effective}
Luong, M.-T.; Pham, H.; and Manning, C.~D.
\newblock 2015.
\newblock Effective approaches to attention-based neural machine translation.
\newblock {\em arXiv preprint arXiv:1508.04025}.

\bibitem[\protect\citeauthoryear{Ma \bgroup et al\mbox.\egroup
  }{2017}]{ma2017improving}
Ma, S.; Sun, X.; Xu, J.; Wang, H.; Li, W.; and Su, Q.
\newblock 2017.
\newblock Improving semantic relevance for sequence-to-sequence learning of
  chinese social media text summarization.
\newblock {\em arXiv preprint arXiv:1706.02459}.

\bibitem[\protect\citeauthoryear{Miller}{1995}]{miller1995wordnet}
Miller, G.~A.
\newblock 1995.
\newblock Wordnet: a lexical database for english.
\newblock {\em Communications of the ACM} 38(11):39--41.

\bibitem[\protect\citeauthoryear{Papineni \bgroup et al\mbox.\egroup
  }{2002}]{Papineni2002BLEU}
Papineni, K.; Roukos, S.; Ward, T.; and Zhu, W.-J.
\newblock 2002.
\newblock Bleu: A method for automatic evaluation of machine translation.
\newblock In {\em Proceedings of the 40th Annual Meeting on Association for
  Computational Linguistics}, ACL '02,  311--318.
\newblock Stroudsburg, PA, USA: Association for Computational Linguistics.

\bibitem[\protect\citeauthoryear{Rajpurkar \bgroup et al\mbox.\egroup
  }{2016}]{RajpurkarZLL16}
Rajpurkar, P.; Zhang, J.; Lopyrev, K.; and Liang, P.
\newblock 2016.
\newblock Squad: 100, 000+ questions for machine comprehension of text.
\newblock {\em CoRR} abs/1606.05250.

\bibitem[\protect\citeauthoryear{RAO~CH and Saha}{2018}]{article}
RAO~CH, D., and Saha, S.~K.
\newblock 2018.
\newblock Automatic multiple choice question generation from text : A survey.
\newblock {\em IEEE Transactions on Learning Technologies} PP:1--1.

\bibitem[\protect\citeauthoryear{See, Liu, and Manning}{2017}]{See_2017}
See, A.; Liu, P.~J.; and Manning, C.~D.
\newblock 2017.
\newblock Get to the point: Summarization with pointer-generator networks.
\newblock {\em Proceedings of the 55th Annual Meeting of the Association for
  Computational Linguistics (Volume 1: Long Papers)}.

\bibitem[\protect\citeauthoryear{Seo \bgroup et al\mbox.\egroup
  }{2016}]{seo2016bidirectional}
Seo, M.; Kembhavi, A.; Farhadi, A.; and Hajishirzi, H.
\newblock 2016.
\newblock Bidirectional attention flow for machine comprehension.
\newblock {\em arXiv preprint arXiv:1611.01603}.

\bibitem[\protect\citeauthoryear{Stasaski and
  Hearst}{2017}]{stasaski2017multiple}
Stasaski, K., and Hearst, M.~A.
\newblock 2017.
\newblock Multiple choice question generation utilizing an ontology.
\newblock In {\em Proceedings of the 12th Workshop on Innovative Use of NLP for
  Building Educational Applications},  303--312.

\bibitem[\protect\citeauthoryear{Sutskever, Vinyals, and
  Le}{2014}]{sutskever2014sequence}
Sutskever, I.; Vinyals, O.; and Le, Q.~V.
\newblock 2014.
\newblock Sequence to sequence learning with neural networks.
\newblock In {\em Advances in neural information processing systems},
  3104--3112.

\bibitem[\protect\citeauthoryear{Tan, Wan, and Xiao}{2017}]{tan2017neural}
Tan, J.; Wan, X.; and Xiao, J.
\newblock 2017.
\newblock From neural sentence summarization to headline generation: A
  coarse-to-fine approach.
\newblock In {\em IJCAI},  4109--4115.

\bibitem[\protect\citeauthoryear{Wang and Chang}{2016}]{wang2016graph}
Wang, W., and Chang, B.
\newblock 2016.
\newblock Graph-based dependency parsing with bidirectional lstm.
\newblock In {\em Proceedings of the 54th Annual Meeting of the Association for
  Computational Linguistics (Volume 1: Long Papers)}, volume~1,  2306--2315.

\bibitem[\protect\citeauthoryear{Xiong, Zhong, and Socher}{2016}]{XiongZS16}
Xiong, C.; Zhong, V.; and Socher, R.
\newblock 2016.
\newblock Dynamic coattention networks for question answering.
\newblock {\em CoRR} abs/1611.01604.

\end{thebibliography}

\end{document}